\documentclass[10pt, a4paper]{article}
\usepackage{lrec}
\usepackage{multibib}
\newcites{languageresource}{Language Resources}
\usepackage{graphicx}
\usepackage{tabularx}
\usepackage{soul}
\usepackage{booktabs}
\usepackage{graphicx}
\usepackage{color}

\usepackage{epstopdf}
\usepackage[latin1]{inputenc}

\usepackage{hyperref}
\usepackage{xstring}
\usepackage{footnote}

\makesavenoteenv{table}

\newcommand{\eg}{\textit{e.g.\,}}

\newcommand{\etc}{\textit{etc.\,}}

\newcommand{\fturl}[1]{\footnote{{\scriptsize \url{#1}}}}

\newcommand*{\thead}[1]{\multicolumn{1}{c}{\bfseries #1}}
\newcommand*{\tsubhead}[1]{\multicolumn{1}{c}{\textit{#1}}}

\title{The IIT Bombay English-Hindi Parallel Corpus}

\name{Anoop Kunchukuttan$^\dagger$, Pratik Mehta\footnotemark[1]$^{1\ddagger}$, Pushpak Bhattacharyya$^\dagger$}

\address{$^\dagger$Center for Indian Language Technology, \\
         Department of Computer Science and Technology, \\
         Indian Institute of Technology Bombay. \\
         $^\ddagger$ College of Information and Computer Sciences, \\
          University of Massachusetts Amherst.  \\
         \{anoopk,pb\}@cse.iitb.ac.in, psmehta@cs.umass.edu \\}

\abstract{
 We present the IIT Bombay English-Hindi Parallel Corpus. The corpus is a compilation of parallel corpora previously available in the public domain as well as new parallel corpora we collected. The corpus contains 1.49 million parallel segments, of which 694k segments were not previously available in the public domain. The corpus has been pre-processed for machine translation, and we report baseline phrase-based SMT and NMT translation results on this corpus. This corpus has been used in two editions of shared tasks at the Workshop on Asian Language Translation (2016 and 2017). The corpus is freely available for non-commercial research. To the best of our knowledge, this is the largest publicly available English-Hindi parallel corpus. 
 \\ \newline  
}
\begin{document}

\maketitleabstract

\footnotetext[1]{work done at IIT Bombay}

\setcounter{footnote}{1}

\section{Introduction}

Hindi is one of the major languages of the world, spoken primarily in the Indian subcontinent, and is a recognised regional language in Mauritius, Trinidad and Tobago, Guyana, and Suriname. In addition, it serves as a major \textit{lingua franca} in India. According to the 2001 Census of India, Hindi has 422 million native speakers and more than 500 million total speakers \cite{wiki:hindi}. It is also an official language of the Union Government of India as well as major Indian states like Uttar Pradesh, Bihar, Rajasthan, \etc and is used for conducting business and administrative tasks. Many languages and dialects in the Gangetic plains are closely related to Hindi \eg Bhojpuri, Awadhi, Maithili, \etc Hindi is the fourth-most spoken language in the world, and third-most spoken language along with Urdu (both are registers of the Hindustani language). In contrast, English is spoken by just around 125 million people in India, of which a very small fraction are native speakers. 

Hence, there is a large requirement for digital communication in Hindi and interfacing with the rest of the word via English. Hence, there is immense potential for English-Hindi machine translation. However, the parallel corpora available in the public domain is quite limited. This work is an effort to consolidate all publicly available parallel corpora for English-Hindi as well as significantly add to the available parallel corpus through corpora collected in the course of this work. 

\section{Dataset}

\begin{table*}
\centering
\begin{tabular}{llr}
\toprule
\textbf{Corpus Id} & \textbf{Source} 			& \textbf{Number of segments}  \\
\midrule
                1 & GNOME          (OPUS) \cite{tiedemann2012parallel}        &                145,706  \\
                2 & KDE4          (OPUS)        &         97,227   \\
                3 & Tanzil        (OPUS)        &        187,080  \\
                4 & Tatoeba       (OPUS)        &        4,698  \\
                5 & OpenSubs2013  (OPUS)        &        4,222  \\
                6 & HindEnCorp    \cite{bojar2014hindencorp}    &         273,885  \\
                7 & Hindi-English Linked Wordnets \cite{bhattacharyya2010indowordnet}      &           175,175  \\
                8 & Mahashabdkosh: Administrative Domain Dictionary$^*$ \cite{kunchukuttan2013transdoop}        &       66,474  \\
                9 & Mahashabdkosh: Administrative Domain Examples$^*$           &     46,825  \\
               10 & Mahashabdkosh: Administrative Domain Definitions$^*$        &        46,523  \\
               11 & TED talks    \cite{abdelali2014amara}   &         42,583  \\
               12 & Indic Multi-parallel corpus \cite{birch2011indic}      &         10,349  \\
               13 & Judicial domain corpus - I$^*$    \cite{kunchukuttan2013transdoop}      &        5,007  \\
               14 & Judicial domain corpus - II$^*$   \cite{kunchukuttan2012experiences}    &         3,727  \\
               15 & Indian Government corpora$^*$     &           123,360  \\
               16 & Wiki Headlines (Provided by CMU: {\scriptsize \url{www.statmt.org/wmt14/wiki-titles.tgz}})       &       32,863  \\
               17 & Gyaan-Nidhi Corpus  $^*$ &                227,123  \\
                  & ({\scriptsize \url{tdil-dc.in/index.php?option=com_download&task=showresourceDetails&toolid=281}}) & \\
\midrule               
               \multicolumn{2}{l}{\textbf{Total}}& \textbf{1,492,827} \\
\bottomrule
\end{tabular}
\caption{Details of the IITB English-Hindi Parallel Corpus (training set). $^*$ indicates new corpora not in the public domain previously.}
\label{tbl:iitb_parallel_stats}
\end{table*}

The parallel corpus has been compiled from a variety of existing sources (primarily OPUS \cite{tiedemann2012parallel}, HindEn \cite{bojar2014hindencorp} and TED \cite{abdelali2014amara}) as well as corpora developed at the \textit{Center for Indian Language Technology}\fturl{www.cfilt.iitb.ac.in} (CFILT), IIT Bombay over the years. The training corpus consists of sentences, phrases as well as dictionary entries, spanning many applications and domains.

\subsection{Corpus Details}

The details of the training corpus are shown in Table \ref{tbl:iitb_parallel_stats}. We briefly describe the new sub-corpora we have added to the collection. For the corpora compiled from existing sources, please refer to the papers mentioned in the table.

\paragraph{Judicial domain corpus - I} contains translations of legal judgements by in-house translators with many years of experience, though not with a legal background. 
\paragraph{Judicial domain corpus - II} contains translation done by graduate students taking a graduate course on natural language processing as part of a course project. This was part of an exercise of collecting translations in complex domain by non-expert translators. The translations included in the corpus were determined to be of good quality by annotators.
\paragraph{Mahashabdkosh} is an online official terminology dictionary website\fturl{e-mahashabdkosh.rb-aai.in}  which is hosted by Department of Official Language, India. It contains Hindi as well as English terms along with definitions and example usage which are translations. The translation pairs were crawled from the website. 
\paragraph{Indian Government corpora} has been  manually collected by CFILT staff from various websites related to the Indian government like the National Portal of India, Reserve Bank of India, Ministry of Human Resource Development, NABARD, \etc
\paragraph{Hindi-English Linked Wordnet} contains bilingual dictionary entries created from the linked Hindi and English wordnets. 
\paragraph{Gyaan-Nidhi Corpus} is a multilingual parallel corpus between English and multiple Indian languages. The data is available in HTML format, hence it is not sentence aligned. We used the sentence alignment technique proposed by \newcite{moore2002fast} to extract parallel corpora from this comparable corpus. This method combines sentence-length models and word-correspondence based models, and requires no language or corpus specific knowledge. We manually checked a small sample of 300 sentences from the parallel sentences extracted. We found that the precision of extraction of parallel sentences was 88.6\%.

\subsection{Corpus Statistics}

The test and dev corpora consist of newswire sentences, which are the same ones as used in the WMT 2014 English-Hindi shared task \cite{bojar2014findings}. 
The training, dev and test corpora consist of 1,492,827 and 520 and 2507 segments respectively. Detailed Statistics are shown in Table \ref{tbl:data-stats}. The Hindi and English OOV rate (for word types) is 11.4\% and 6.7\%. 


\begin{table}
\centering
\resizebox{0.45\textwidth}{!}{%

\begin{tabular}{lcrrr}
\toprule 
               &    \thead{Language}   & \thead{Train} & \thead{Test} & \thead{Dev} \\ 
\midrule                              
\#Sentences     &               &   1,492,827        & 2,507         & 520         \\ 
\midrule   
\#Tokens        &  eng          &  20,667,259        &   57,803           &   10,656      \\ 
			  &   hin          &  22,171,543        &   63,853           &    10,174      \\ 
\midrule   
\#Types         &  eng          &      250,782       &   8,957         &     2,569         \\ 
			  &   hin         &      343,601       &    8,489          &      2,625        \\ 
\bottomrule
\end{tabular}
}

\caption{Statistics of data sets}
\label{tbl:data-stats}
\end{table}

\section{Baseline Systems}

We trained baseline machine translation models using the parallel corpus with popular off-the-shelf machine translation toolkits to provide benchmark translation accuracies for comparison. We trained phrase-based Statistical Machine Translation (PBSMT) systems as well as Neural Machine Translation systems for English-Hindi and Hindi-English translation. 

\subsection{Data Preparation}
\noindent \textbf{Text Normalization:} For Hindi, characters with \textit{nukta} can have two Unicode representations. In one case, the character and nukta are represented as two Unicode characters. In the other case, a single Unicode character represents the composite character. We choose the former representation. The normalization script is part of the \textit{IndicNLP}\fturl{anoopkunchukuttan.github.io/indic_nlp_library}  library . 

For English, we used true-cased representation for our experiments. However, the parallel corpus being distributed is available in the original case. 

\noindent \textbf{Tokenization:} We use the \textit{Moses} tokenizer for English and the \textit{IndicNLP} tokenizer for Hindi.

\subsection{SMT Setup}

We trained PBSMT systems with \textit{Moses}\fturl{www.statmt.org/moses} \cite{koehn2007moses}. We used the \textit{grow-diag-final-and} heuristic for extracting phrases, lexicalised reordering  and Batch MIRA \cite{cherry2012batch} for tuning (default parameters). We trained 5-gram language models with Kneser-Ney smoothing using \textit{KenLM} \cite{heafield2011kenlm}. We used the \textit{HindMono} \cite{bojar2014hindencorp} corpus for Hindi and the WMT NEWS Crawl 2015 corpus for English as additional monolingual corpora to train language models. These contain roughly 44 million and 23 million sentence for Hindi and English respectively.

\subsection{NMT Setup}

We trained a subword-level encoder-decoder architecture based NMT system with attention \cite{bahdanau2015neural}. We used \textit{Nematus} \fturl{github.com/EdinburghNLP/nematus} \cite{sennrich2017nematus} for training our NMT systems.

\noindent \textbf{Vocabulary:} We used Byte Pair Encoding (BPE) to learn the vocabulary (with 15500 merge operations) \cite{sennrich2016neural}.  We used the \textit{subword-nmt} \fturl{github.com/rsennrich/subword-nmt} tool for learning the BPE vocabulary.  Since the writing systems and vocabularies of English and Hindi are separate, BPE models are trained separately.

\noindent \textbf{Network parameters:} The network contains a single hidden encoder and decoder RNN layer, containing 512 GRU units each. The dimension of input and output embedding layers is 256 units. 

\noindent \textbf{Training details:} The model is trained with a batch size of 50 sentences and maximum sentence length of 100 using Adam optimizer \cite{kingma2014adam} with a learning rate of 0.0001. The output parameters were saved after every 10,000 iterations. We used early-stopping based on validation loss with \textit{patience=10}.

\noindent \textbf{Decoding:} We used a beam size of 12. We decoded the test set with an ensemble of four models (best model and the last three saved models). 

\subsection{Results}

We evaluated our system using BLEU \cite{papineni2002bleu} and METEOR \cite{banerjee2005meteor}. We used a \textit{METEOR-Indic}\fturl{github.com/anoopkunchukuttan/meteor_indic}, a  customized version of METEOR Indic, for evaluation of Hindi as target language. \textit{METEOR-Indic} can perform synonym matches for Indian languages using synsets from IndoWordNet \cite{bhattacharyya2010indowordnet}. It can also perform stem matches for Indian languages using a trie-based stemmer \cite{bhattacharyya2014facilitating}. This is useful for a morphologically rich language like Hindi.

Table \ref{tbl:results} shows the results of our experiments. 

\begin{table}
\centering
\begin{tabular}{lrrrr}
\toprule 
\thead{System} & \multicolumn{2}{c}{\textbf{eng-hin}} & \multicolumn{2}{c}{\textbf{hin-eng}} \\
\cmidrule(lr){2-3}
\cmidrule(lr){4-5}
 & \tsubhead{BLEU} & \tsubhead{METEOR} & \tsubhead{BLEU} & \tsubhead{METEOR} \\
\midrule
SMT & 11.75 & 0.313 & 14.49 & 0.266 \\
NMT & 12.23 & 0.308 & 12.83 & 0.219 \\
\bottomrule
\end{tabular}
\caption{Results for Baseline Systems}
\label{tbl:results}
\end{table}


\section{Availability}

The homepage for the dataset can be accessed here: \url{http://www.cfilt.iitb.ac.in/iitb_parallel}. The new corpora we release are available for research and non-commercial use under a Creative Commons Attribution-NonCommercial-ShareAlike License \fturl{http://creativecommons.org/licenses/by-nc-sa/4.0}. The corpora we compiled from other sources are available under their respective licenses. The sub-corpora (in the corpus distribution that we make available) are in the same order as listed in the Table \ref{tbl:iitb_parallel_stats}, so they can be separately extracted, if required (\eg for domain adaptation). 

\section{Conclusion and Future Work}

We presented the IIT Bombay English-Hindi Parallel corpus version 1.0, and provided benchmark baseline SMT and NMT results on this corpus. This corpus has been used for the two shared tasks (Workshop on Asian Language Translation 2016 and 2017). The \textit{HindiEn} component of the corpus has also been used for the WMT 2014 shared task. The corpus is available under a Creative Commons Licence. 

In future, we plan to enhance the corpus from additional sources, mostly websites of the Government of India which is still a largely untapped source of parallel corpora.  We also plan to build stronger baselines like pre-ordering with PBSMT \cite{AnanthakrishnanJayprasadHegde2008} for English-Hindi translation, and use of synthetic corpora generated via back-translation for NMT systems \cite{sennrich2016edinburgh}. 

\section{Acknowledgements}
We thank past and present members of the Center for Indian Language Technology for their efforts in creating various parts of the corpora over the years: Pallabh Bhattacharjee, Kashyap Popat, Rahul Sharnagat, Mitesh Khapra, Jaya Jha, Rajita Shukla, Laxmi Kashyap, Gajanan Rane and many members of the Hindi WordNet team. 
We also thank the Technology Development for Indian Languages (TDIL) Programme and the Department of Electronics \& Information Technology, Government of India for their support.

\section{Bibliographical References}
\label{main:ref}

\bibliographystyle{lrec}
\bibliography{lrec2018_iitbparallel}


\end{document}